\documentclass[letterpaper]{article}
\usepackage{aaai}
\nocopyright
\usepackage{times}
\usepackage{helvet}
\usepackage{courier}

\usepackage{amssymb}
\usepackage{amsfonts}
\usepackage{amstext}
\usepackage{amsmath}

\RequirePackage{amsthm}
\RequirePackage{amsmath}
\RequirePackage{amssymb}
\theoremstyle{plain}

\theoremstyle{definition}
\newtheoremstyle{bfnote}
{}                {}                {\slshape}        {}                {\bf}       {.}               { }               {\thmname{#1}\thmnumber{ #2}\thmnote{ (#3)}}       \theoremstyle{bfnote}
\newtheorem{definition}{Definition}

\usepackage{algorithm}
\usepackage{algorithmicx}
\usepackage[noend]{algpseudocode}
\renewcommand*\Call[2]{\textproc{#1}(#2)}

\newcommand\NoThen{\renewcommand\algorithmicthen{}}

\algnewcommand{\IfThen}[2]{\State \algorithmicif\ #1\ \textbf{then}\ #2}
\algnewcommand{\IfThenElse}[3]{\State \algorithmicif\ #1\ \textbf{then}\ #2\ \algorithmicelse\ #3}

\algrenewcommand\algorithmicindent{0.9em}

\usepackage{todonotes}
\newcommand{\frm}[2][noinline]{\todo[author=Felipe,size=\tiny,color=red!60,#1]{{#2}}\ }
\newcommand{\mau}[2][noinline]{\todo[#1,color=yellow!80,size=\tiny]{MAU: #2}}
\usepackage[normalem]{ulem}

\usepackage{listings}
\lstdefinelanguage{PDDL}
{
  sensitive=false,    morecomment=[l]{;}, alsoletter={:,-},   morekeywords={
    define,domain,problem,not,and,or,when,forall,exists,either,
    :domain,:requirements,:types,:objects,:constants,
    :predicates,:action,:parameters,:precondition,:effect,
    :fluents,:primary-effect,:side-effect,:init,:goal,
    :strips,:adl,:equality,:typing,:conditional-effects,
    :negative-preconditions,:disjunctive-preconditions,
    :existential-preconditions,:universal-preconditions,:quantified-preconditions,
    :functions,assign,increase,decrease,scale-up,scale-down,
    :metric,minimize,maximize,
    :durative-actions,:duration-inequalities,:continuous-effects,
    :durative-action,:duration,:condition
  }
} \lstdefinelanguage{JSHOP}
{
  sensitive=false,    morecomment=[l]{;}, alsoletter={:,-},   morekeywords={
    defdomain,defproblem,not,and,or,imply,forall,assign,call,nil,
    :first,:sort-by,:immediate,:unordered,:operator,:method,:protection,:-
  }
} \lstdefinelanguage{HDDL}
{
  sensitive=false,    morecomment=[l]{;}, alsoletter={:,-},   morekeywords={
    define,domain,problem,not,and,or,forall,exists,
    :domain,:requirements,:typing,:hierarchy,:equality,:negative-preconditions,
    :method-precondition,:universal-preconditions,:universal-effects,:existential-preconditions,
    :types,:predicates,:action,:parameters,:precondition,:effect,
    :task,:method,:subtasks,:tasks,:ordered-subtasks,:ordered-tasks,:ordering,:constraints,
    :htn,:init,:goal,:objects
  }
}

\begin{document}

\title{HyperTensioN and Total-order Forward Decomposition optimizations}

\author{
Maurício Cecílio Magnaguagno$^1$, Felipe Meneguzzi$^2$ \textnormal{and}\ Lavindra de Silva$^3$\\
$^1$Independent researcher\\
$^2$University of Aberdeen, Aberdeen, UK\\
$^3$Department of Engineering, University of Cambridge, Cambridge, UK\\
\texttt{maumagnaguagno@gmail.com}\\ 
\texttt{felipe.meneguzzi@abdn.ac.uk}\\
\texttt{lavindra.desilva@eng.cam.ac.uk}
}

\maketitle

\begin{abstract}
\begin{quote}
Hierarchical Task Networks (HTN) planners generate plans using a decomposition process with extra domain knowledge to guide search towards a planning task.
While domain experts develop HTN descriptions, they may repeatedly describe the same preconditions, or methods that are rarely used or possible to be decomposed.
By leveraging a three-stage compiler design we can easily support more language descriptions and preprocessing optimizations that when chained can greatly improve runtime efficiency in such domains.
In this paper we evaluate such optimizations with the HyperTensioN HTN planner, used in the HTN IPC 2020.
\end{quote}
\end{abstract}

\section{Introduction}

Hierarchical planning was originally developed as a means to allow planning algorithms to incorporate domain knowledge into the search engine using an intuitive formalism~\cite{nau1999shop}.
Hierarchical Task Network (HTN) is the most widely used formalism for hierarchical planning, having been implemented in a variety of systems rendered in different (though conceptually similar) input languages~\cite{de2015hatp,nau2003shop2,Ilghami2003}. 
Recent research has re-energized work on HTN planning formalisms and search procedures, leading to a new generation of HTN planners~\cite{Bercher:2017ui,Holler:2018tp,Hoeller2020HDDL,holler2020htn}. 
In this paper, we outline key design elements, features, and optimizations of the HyperTensioN planner, as submitted to the 2020 International Planning Competition (IPC)\footnote{ipc-2020.hierarchical-task.net}. 
Specifically, we focus on the compilation of HTN instances into Ruby/C++ programs, as well as the optimizations based on transformation of HTN domains and problems to minimize backtracking.

\section{Three-stage design architecture}

HyperTensioN was originally developed to automatically convert classical planning instances to hierarchical planning instances~\cite{magnaguagno2017method}.
This required at least a PDDL~\cite{mcdermott1998pddl} parser (front-end) and a (J)SHOP~\cite{Ilghami2003} description compiler (back-end).
By keeping front-end and back-end separate it was also possible to add a Ruby compiler to generate code compatible with our implementation of a lifted Total-order Forward Decomposition (TFD)~\cite[chapter~11]{ghallab2004automated} planner.
This compilation approach is very similar to that in JSHOP~\cite{Ilghami2003}.
Parser and compiler modules use the same Intermediate Representation (IR) to share planning instance data, which middle-end extensions can further process. 
Extensions fill gaps between description languages, analyze or optimize descriptions, independent of the target planner, input and output language. 

\begin{figure}[!t]
\centering
\includegraphics[width=0.48\textwidth]{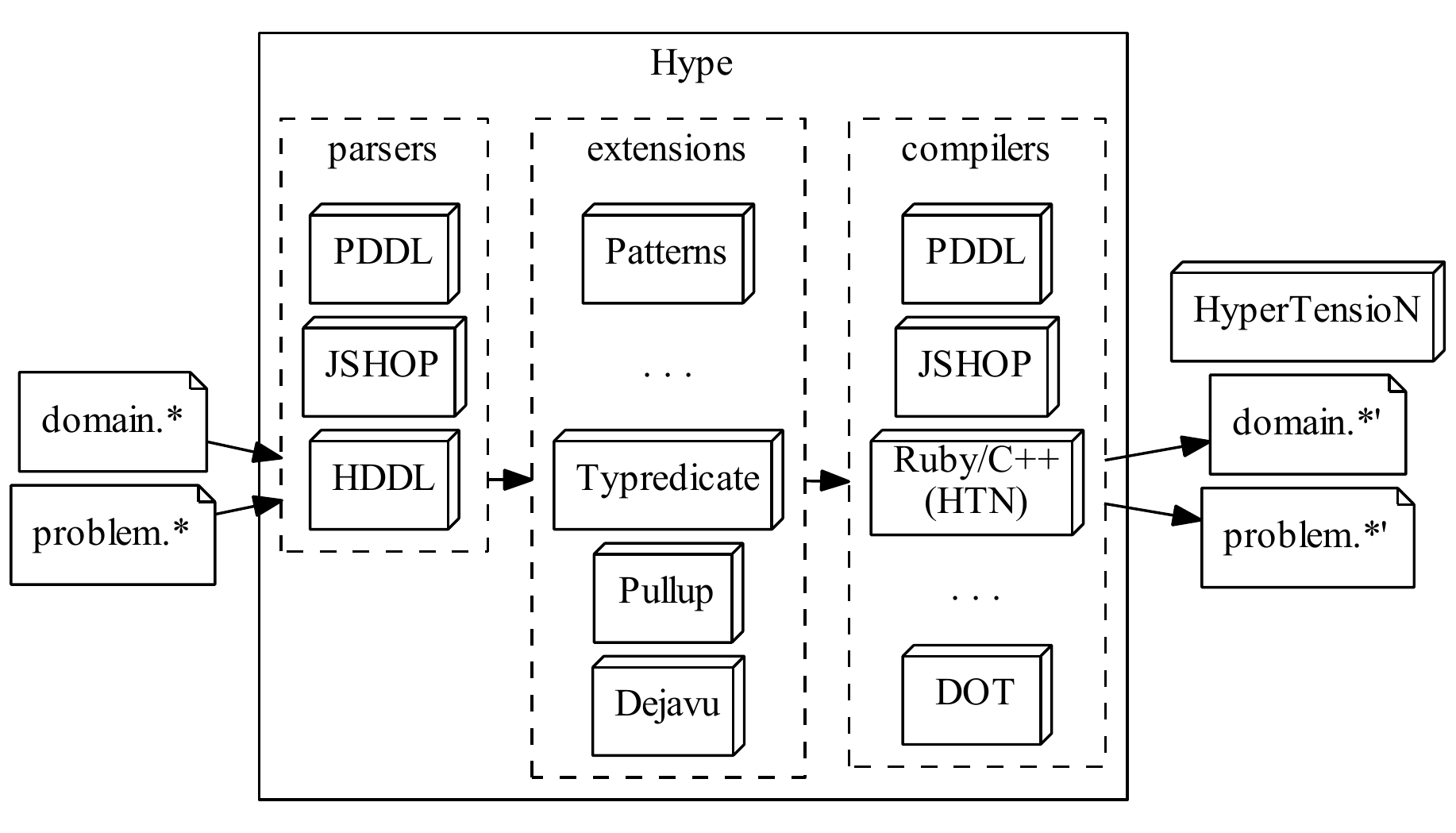}
\caption{Hype acts as a three-stage compiler before linking with the HyperTensioN planner.}
\label{fig:hype}
\end{figure}

This level of flexibility facilitates developing support for new languages, while remaining compatible with the already available extensions. For example the DOT~\cite{ellson2001graphviz} compiler for debugging and the HDDL~\cite{Hoeller2020HDDL} parser for the IPC.
As the project grew, the three-stage compiler and the TFD planner modules split in two, as shown in Fig.~\ref{fig:hype}.
The Hype tool controls module execution at each stage, allowing multiple middle-ends to run, even repeatedly, before compilation into the target representation.
The HyperTensioN TFD planner completes the HTN compiler output to finish this pipeline with the plan output.
Eventually, we extended the core HyperTensioN search procedure to a variety of other planning tasks, including search on hybrid symbolic-numeric domains~\cite{Magnaguagno2020}.

\section{Background}

Hierarchical planning focuses on solving planning tasks while including and exploiting human knowledge about problem decomposition using a hierarchy of decisions and operators as the planning domain.
The quality of this hierarchy of decisions greatly impacts the planner performance and resulting plan.
The elements of classical planning are here defined and used as a base for the hierarchical elements. 

A classical planning problem instance is defined by initial and goal states that encode properties of the objects in the world at a particular time.
In order to achieve the desired goal one must respect the rules of the domain, which limit which transitions are valid.
Such transitions are the domain operators and are defined as preconditions and effects.
Preconditions and effects use predicates and free variables that, when unified with the available objects, enumerate the possible actions to be performed.
During the planning process, states are tested to check which actions are applicable based on their preconditions being satisfied.
For the applicable actions a new possible state is created by applying its effects.
Preconditions are satisfied when the constraint formula (usually a conjunction of predicates) is valid at the current state the action is being applied.
The effects contain positive and negative sets of predicates to be added and removed by actions, respectively, changing object properties of the current state.
Once a state that satisfies the goal is reached, the path taken is the plan or solution, a finite sequence of actions~\cite{nebel2011compilability}.

Classical planning formalisms comprise the following elements:

\begin{definition}[Terms]\label{def:term}
Terms are symbols that represent objects or variables.
We call \textbf{O} the finite set of objects available.
\end{definition}

Predicates represent object relations that can be observed or changed during planning.
Predicates can also be seen as constraints between terms.
When all terms of a predicate are objects we call it a ground predicate, otherwise the predicate contain at least one variable term and is a lifted predicate.
Ground predicates are obtained from lifted predicates through Unification, replacing variable terms with objects.

\begin{definition}[Predicates]\label{def:predicate}
Predicates are defined by a signature \textbf{name} applied to a sequence of N \textbf{terms}, represented by t$_n$, \textbf{p} = \textbf{$\langle$name(p), terms(t$_1$, ..., t$_n$)$\rangle$}.
We call \textbf{F} the finite set of facts, comprised of all ground predicates.
\end{definition}

\begin{definition}[Unification]\label{def:unification}
When a predicate \textbf{p} have all its variable terms, variables(\textbf{p}) = t $\in$ terms(\textbf{p}) $\land$ t $\notin$ \textbf{O}, replaced by \textbf{O} objects, we have unified a lifted predicate to a ground one.
One can enumerate several substitutions using the cartesian product between variable terms and \textbf{O}, such that variables(\textbf{p}) x O = \{(v, o) | v $\in$ variables(\textbf{p}) $\land$ o $\in$ O\}.
Then \{(v$_0$, o$_0$), (v$_0$, o$_1$), ..., (v$_n$, o$_n$)\} is the set of replacements of \textbf{p} to obtain ground predicates.
\end{definition}

A state is a finite set of ground predicates that describes a world configuration at a particular time.
Partial states may be used to represent only what we are interested in, in a closed-world assumption where we have full observability.
Partial states may also be used to represent only predicates whose state is certainly know, in an open-world assumption, where we may lack certainty about which predicates are true or false.

\begin{definition}[State]\label{def:state}
States are represented by \textbf{S} = $\langle$p$_1$, ..., p$_n\rangle$, a set of ground predicates.
\end{definition}

States can be modified respecting constraints that describe when a modification is \textsc{applicable}, and the modification itself, described by \textsc{apply}.

\begin{definition}[Applicable]\label{def:applicable}
A set of ground predicates is considered applicable when its positive part is contained in the current state, while the negative part is disjointed, described by the function $\textsc{applicable}(pre, State) : pre^+ \subseteq State \land pre^- \cap State = \varnothing$.
\end{definition}

\begin{definition}[Apply]\label{def:apply}
Apply is a function that removes and adds distinct effect sets of ground predicates to create a new State from the current State, described by the function $\textsc{apply}(a, State) : (State~-~$eff$^-$(a)) $\cup$ eff$^+$(a).
\end{definition}

Each set of predicates to be used in \textsc{applicable} and \textsc{apply} calls can be generalized, using variable terms, to make an Operator.
More complex preconditions and effects consider expressions, quantifiers and conditions instead of just set operations to obtain more expressive Operators.
The Operators can be Unified with \textbf{O} to obtain the full set of possible Actions. 

\begin{definition}[Operator]\label{def:operator}
Operators are represented by a 4-tuple \textbf{o} = \textbf{$\langle$name(o), pre(o), eff(o), cost(o)$\rangle$}:\textbf{name(o)} is the description or signature of \textbf{o};
\textbf{pre(o)} are the preconditions of \textbf{o};
The preconditions contain positive and negative sets, \textbf{pre$^+$(o)} and \textbf{pre$^-$(o)}, of predicates that must be applicable in the current state for action \textbf{o} to be applied;
\textbf{eff(o)} are the effects of \textbf{o};
The effects contain positive and negative sets, \textbf{eff$^+$(o)} and \textbf{eff$^-$(o)}, that add and remove predicates from the state, respectively;
\textbf{cost(o)} represents the cost of applying this operator, usually one or zero.
The finite set of operators is called \textbf{Op}.
The predicates that are present in the effects of \textbf{Op} are called fluent predicates, while the ones not present are called rigid predicates, which makes rigid predicates the same for every State.
Rigid predicate preconditions with a single term may be declared as types to improve readability through a hierarchical structure, and to be exploited by unification.
\end{definition}

\begin{definition}[Action]\label{def:action}
Actions are instantiated/ground operators obtained from the Unification process of Definition~\ref{def:unification}.
During the planning process, each action \textit{a} that is $\textsc{applicable}(pre(a), State_n)$ can create a new reachable State, $State_{n+1} \gets \textsc{apply}(a, State_n)$.
The finite set of actions available is called \textbf{A}.
\end{definition}

Classical Planning is goal-driven, which requires the description of a Initial state and a Goal state to plan for.
The planner is responsible for finding a Plan, a sequence of Actions from \textbf{A}, that when applied to the Initial state will satisfy the Goal state description.

\begin{definition}[Initial state]\label{def:initialstate}
The Initial state is a complete State, in a closed-world, represented by \textbf{I} $\subseteq$ \textbf{F}, which is defined by a set of predicates that represent the current state of the environment.
\end{definition}

\begin{definition}[Goal state]\label{def:goalstate}
Goal state is a partial State represented by \textbf{G} $\subseteq$ \textbf{F}, which is defined by a set of predicates that we desire to achieve by applying the actions available. 
\end{definition}

Each Planning Instance is made of a generic domain \textbf{D} and a specific situation within this domain to be solved, described by \textbf{I} and \textbf{G}.
The solution is a Plan.
Not all Planning instances can be solved, as some \textbf{G} may be unreachable based on \textbf{I} and \textbf{A}, which results in planning failure.

\begin{definition}[Domain]\label{def:domain}
Domain brings all problem independent elements together in the tuple \textbf{D}~=~\textbf{$\langle$F,~A$\rangle$}.
\end{definition}

\begin{definition}[Plan]\label{def:plan}
Plan is the solution concept of a planning problem and is represented by a sequence of actions that when applied in a specific order will modify \textbf{I} to \textbf{G} in \textbf{D}, \textbf{$\pi$} = \textbf{$\langle$ a$_1$, ..., a$_n\rangle$}.
An empty plan solves \textbf{G} $\subseteq$ \textbf{I}.
\end{definition}

\begin{definition}[Planning instance]\label{def:instance}
Planning instance represented by the 3-tuple \textbf{P} = \textbf{$\langle$D, I, G$\rangle$}, planners take as input \textbf{P} and return either $\pi$ or \textit{failure}.
\end{definition}

With domain knowledge one knows which action sequences are frequently used to solve subproblems in specific domains~\cite{ponsen2005automatically}.
To exploit such domain knowledge about problem decomposition we shift from goal states to tasks.
The goal state is implicitly achieved by the plan obtained from the tasks, just as a cake from a recipe.
The ingredients available act as decision points, that can be used to create a hierarchy of decisions, while preferences appear as the order in which such decisions are considered.

The problem for hierarchical planning is defined with initial state and tasks.
Each task corresponds to a starting node in a hierarchy, which comprises two types of nodes: primitive tasks that map directly to an operator; and non-primitive tasks that select a method that decomposes to subtasks.
Applicable tasks are refined into subtasks until only primitive tasks remain.

\begin{definition}[Task]\label{def:task}
A task is represented by a signature \textbf{name(task)} applied to a sequence of N \textbf{terms} that act as parameters, forwarding ground values to be used by the task, \textbf{task} = \textbf{$\langle$name(task), terms(t$_1$, ..., t$_n$)$\rangle$}.
\end{definition}

A set of tasks to be decomposed by an instance is called \textbf{T}.
During each step of the planning process the first task is removed from \textbf{T} by \textsc{shift}, and mapped by name to an operator (primitive task), equivalent to the one from Definition~\ref{def:operator}, or method (non-primitive/abstract task).

\begin{definition}[Method]\label{def:method}
A method is a 3-tuple \textbf{m} = \textbf{$\langle$name(m), pre(m), tasks(m), constr(m)$\rangle$}, where:
\textbf{name(m)} is the description or signature of \textbf{m};
\textbf{pre(m)} are the preconditions of \textbf{m};
\textbf{tasks(m)} are the subtasks of \textbf{m}, replacing the original task for new tasks;
\textbf{constr(m)} are the ordering constraints imposed to the subtasks of \textbf{m}.
Each ordering constraint describes the relation between two subtasks, e.g. $t_i \prec t_j$.
In methods where not all subtasks are ordered, the planner is free to find an ordering that achieves the plan.
The finite set of methods available is called \textbf{M}.
\end{definition}

During the HTN planning process, each possible \textsc{decomposition} of \textbf{m} is found by searching which methods match the current task, \textbf{name(t)} = \textbf{name(m)} for t $\in$ \Call{shift}{\textbf{T}} $\land$ \textbf{m} $\in$ \textbf{M}.
In this work we will limit to total-order decomposition, \textbf{constr(m)} = $t_{n-1} \prec t_n$ for $n$ = |\textbf{tasks(m)}| and $n \geq 2$, which simplifies \Call{shift}{\textbf{T}} to consider only the first task.
Partial ordering requires bookkeeping of the ordering constraints to interleave tasks, and more complex precondition descriptions, as tasks can be accomplished in many ways not described by total ordering.
The preconditions of the first task to be decomposed, \textbf{pre}(\Call{shift}{\textbf{T}}), have their variable terms unified to objects based on the current state.
\textbf{G} may be empty in hierarchical planning, however, when provided it can be verified after planning or compiled into a goal operator with the partial state as preconditions, and empty effects.
A task that maps to the goal operator is the last task of \textbf{T}.

\begin{definition}[HTN domain]\label{def:htndomain}
The classical planning domain is extended with \textbf{M} methods to make the HTN domain.
The HTN domain is represented by \textbf{D} = \textbf{$\langle$F, A, M$\rangle$}.
\end{definition}

\begin{definition}[HTN plan]\label{def:htnplan}
HTN plan represented by a sequence of actions that when applied in a specific order will modify \textbf{I} to an implicit \textbf{G} defined by \textbf{T}, $\pi$ = \textbf{$\langle$a$_1$, ..., a$_n\rangle$}.
\end{definition}

\begin{definition}[HTN planning instance]\label{def:htninstance}
HTN planning instance represented by the 3-tuple \textbf{P} = \textbf{$\langle$D, I, T$\rangle$} and returns $\pi$ or \textit{failure}.
\end{definition}

\subsection{Total Forward Decomposition with stack limit}

HyperTensioN is based on Total-order Forward Decomposition (TFD)~\cite[chapter~11]{ghallab2004automated}, presented in Algorithm~\ref{alg:TFD}.
It contains one important modification, it backtracks as the stack limit is reached, similar to depth limited search.
Several domains contain a recursive description that may consume the whole stack without extra mechanisms to avoid decomposing the same task for the same state.
The rest of the algorithm remains the same, decomposing the next task as primitive or non-primitive.
Applicable primitive tasks are applied, while non-primitive tasks may contain multiple related methods that, when applicable, add subtasks to be decomposed before calling the function recursively.
The process continues until failure is returned or no more tasks are left to be decomposed, building the plan while returning from each function call.

\begin{algorithm}[h]
  \caption{Total-order Forward Decomposition planner}
  \label{alg:TFD}
  \begin{algorithmic}[1]
  \NoThen
    \Function{TFD}{\textbf{S}, \textbf{T}, \textbf{D}}
    \IfThen {\textbf{T} = $\varnothing$}{\Return empty $\pi$} \label{tfd-base}
    \IfThen{stack limit reached}{\Return \textit{failure}}
    \State t $\gets$ \Call{shift}{\textbf{T}} \label{tfd-shift}
    \If {t is a primitive task}
      \For {$t_{applicable}$ $\in$ \Call{applicable}{\textbf{pre}(t), \textbf{S}}}
        \State $\pi$ $\gets$ \Call{TFD}{\Call{apply}{$t_{applicable}$, \textbf{S}}, \textbf{T}, \textbf{D}}
        \IfThen {$\pi$ $\neq$ \textit{failure}}{\Return $t_{applicable}$ $\cdot$ $\pi$}
      \EndFor
    \ElsIf {t is a non-primitive task}
      \For {m $\in$ \Call{decomposition}{t, D}}
        \For {\textbf{tasks(}m\textbf{)} $\in$ \Call{applicable}{\textbf{pre}(m), \textbf{S}}}
          \State $\pi$ $\gets$ \Call{TFD}{\textbf{S}, \textbf{tasks}(m) $\cdot$ \textbf{T}, \textbf{D}}
          \IfThen {$\pi$ $\neq$ \textit{failure}}{\Return $\pi$}
        \EndFor
      \EndFor
    \EndIf
    \State \Return \textit{failure}
    \EndFunction
  \end{algorithmic}
\end{algorithm}

\section{Domain transformation}

To improve planning speed the compiler was optimized to compress the state structure by removing rigid predicates and treating them as ``constant information''. More importantly, we developed extensions to improve the IR to support: (1) better unification exploiting type information; (2) early testing
of rigid parts of method/action preconditions during decomposition; and (3) a cycle detection mechanism.

\subsection{Typredicate}

\begin{figure}[b]
\centering
\includegraphics[width=0.3\textwidth]{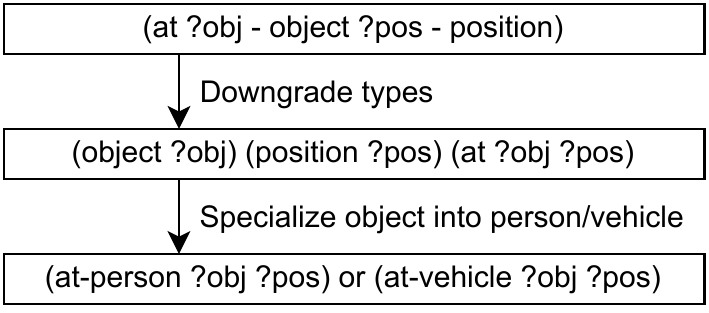}
\caption{Typredicate specializes \textit{at} into \textit{at-person} and \textit{at-vehicle} predicates to consider smaller sets of possible values.}
\label{fig:typredicate}
\end{figure}

This extension involves constraining the substitutions attempted on variables occurring in predicates, by making better use of constant/parameter types (if the domain expert has not already done so).
For example, suppose the predicate \textit{(at ?obj -- object ?pos -- position)} is defined in the domain, which is used in the action \textit{(move ?obj -- vehicle ?pos -- position)} to both check and update a vehicle’s position. 
Suppose also that we are given the following type hierarchy: ``\textit{person vehicle position -- object}''. 
Then, though the \textit{move} action will never require nor modify the position of a person or position, the \textit{?obj} variable occurring in the precondition of the action may still be substituted by constants of type \textit{person} and \textit{position}, as \textit{?obj} is defined in the predicate to be of the parent type \textit{object}. 
Since constants of each type are mutually exclusive by virtue of being subtypes of the same parent type, we  preclude such substitutions by specializing \textit{(at ?obj -- object ?pos -- position)} into predicates \textit{(at-vehicle ?obj -- vehicle ?pos -- position)} and \textit{(at-person ?obj -- person ?pos -- position)}, and replacing occurrences of \textit{(at ?obj ?pos)} with the appropriate specialized predicates in the domain, initial and goal states.
This example is presented in Figure~\ref{fig:typredicate}.
Typredicate currently only specializes predicates to the leaves of the type hierarchy, but it can be straightforwardly extended to specialize to intermediate levels.
Typredicate is not limited to typed domains, as it can infer types based on unary rigid predicates contained in  preconditions, e.g. \textit{(person ?obj)}.
By specializing predicates we make planning more efficient, as unification uses smaller (disjoint) sets of objects, i.e., without extraneous objects, while also making  the  Pullup extension more ``complete''.

Typredicate, Algorithm~\ref{alg:typredicate}, starts selecting rigid predicates with a single term within the positive preconditions of operators.
Such predicates are equivalent to types, \textit{(type ?term)}, and since our parsers always downgrade types to rigid preconditions, the algorithm does not distinguish between them.
Terms that have only one related type are ignore, as it cannot be split into smaller subsets.
Predicates related to supertypes are also currently ignored, as most domains use few supertypes compared to the complexity added to consider them.
Otherwise find the possible subsets and create the new predicates.
The last step requires modifying all elements in which the new predicates may appear, such as operators, methods, initial and goal state.

\begin{algorithm}[!ht]
  \caption{Typredicate}
  \label{alg:typredicate}
  \begin{algorithmic}[1]
  \NoThen
\Procedure{Typredicate}{\textbf{Op}, \textbf{M}, \textbf{I}, \textbf{G}}
      \State new\_predicates = table
      \State operator\_types = table
      \For{o $\in$ \textbf{Op}}
        \State t = operator\_types[\textbf{name}(o)]
        \State \Call{match\_types}{o, t}
        \State \Call{find\_types}{\textbf{pre}(o), t, new\_predicates}
        \State \Call{find\_types}{\textbf{eff}(o), t, new\_predicates}
      \EndFor
      \State transf = table
      \For{$\langle$pre, typ$\rangle$ $\in$ new\_predicates}
        \State typ = \Call{unique}{typ} \Comment{remove duplicates}
        \IfThen{$|$typ$|$ = 1}{\textbf{continue}}
\If{typ contains only leaf types} \Comment{No supertypes}
          \For{t $\in$ types, $\forall$p $\in$ t, p $\in$ \textbf{I}}
            \State transf[pre, t] = \Call{join}{pre, t, '\_'} \EndFor
        \EndIf
      \EndFor
      \IfThen{transf = $\varnothing$}{\Return}
      \For{o $\in$ \textbf{Op}}
        \State t = operator\_types[\textbf{name}(o)]
        \State \Call{replace\_predicates}{\textbf{pre}(o), t, transf}
        \State \Call{replace\_predicates}{\textbf{eff}(o), t, transf}
      \EndFor
      \For{m $\in$ \textbf{M}}
        \State t = table
        \State \Call{match\_types}{m, t}
        \State \Call{replace\_predicates}{\textbf{pre}(m), t, transf}
      \EndFor
      \State \Call{ground\_transf}{\textbf{I}, transf}
      \State \Call{ground\_transf}{\textbf{G}, transf}
    \EndProcedure

    \Procedure{match\_types}{elemt, types}
      \For{$\langle$pre, terms$\rangle$ $\in$ \textbf{pre$^+$}(elemt)}
        \IfThen{pre is rigid $\land$ $|$terms$|$ = 1}{types[terms] = pre}
      \EndFor
    \EndProcedure

    \Procedure{find\_types}{group, types, new\_predicates}
      \For{$\langle$pre, terms$\rangle$ $\in$ group}
        \State values = types of terms
        \State \Call{insert}{new\_predicates[pre], values}
      \EndFor
    \EndProcedure

    \Procedure{ground\_transf}{group, transf}
      \For{$\langle$pre, terms$\rangle$ $\in$ group}
        \State new\_pre = transf[pre, terms]
        \IfThen{new\_pre $\neq$ $\varnothing$}{\Call{replace}{pre, new\_pre}}
      \EndFor
    \EndProcedure
  \end{algorithmic}
\end{algorithm}

\subsection{Pullup}

The Pullup extension implements the main optimization technique that underpins HyperTensioN's performance by ``pulling up'' preconditions in the hierarchy.
A literal in the precondition (which is a conjunction/set of literals) of an action occurring in a method is added (after variable substitutions) to the precondition of the method if the literal is not possibly brought about by an earlier step in the method, i.e., any solution for the method will require the literal to hold at the start; a literal is deemed to be possibly brought about (cf. ``mentioned'' \cite{deSilva-2016}) by a step if there is a literal asserted by an action yielded by the step s.t. the two literals have the same predicate symbol.\footnote{We also implemented a stronger notion, closer to that of ``mentioned'', but saw no improvement w.r.t. the sample IPC domains.} 
We pull up method preconditions as follows. 
A (possibly pulled up) literal in the precondition of a method is deemed to be part of the precondition of the task that is accomplished by the method if the literal is ``locally rigid'', i.e.,  shared by all method preconditions related to the same task.
Given a planning problem, each iteration of the algorithm pulls up literals by one level, considering actions the lowest level, and the algorithm terminates when it reaches a fixed point--when no literals ``moved'' in the previous iteration.

A literal that is always pulled up from an action/method precondition is removed from it, as the literal will be tested earlier in the decomposition.
Moreover, using the planning problem,
literals that are always true (w.r.t. the problem) are removed from preconditions based on the  unifications that are possible,
and actions/methods that contain contradictions in preconditions are removed together with their associated ``branches''. Interestingly,
branch removal may enable pulling up additional literals by exposing ``hidden'' (see \cite{deSilva-2016}) rigid literals.

Consider the abstract domain from Figure~\ref{alg:pullup} as an example.
In this domain the preconditions of Operator1 can be pulled up to Method2, and the preconditions shared with Method3 may be pulled up to Method1.
Preconditions of Operator2 can also be pulled if they are locally rigid, not affected by Operator1 effects, and further if not affected by any path that decomposes Task1.
In some cases all preconditions are pulled up towards Method1, moving the decisions to an earlier stage, while other cases simply remove repeated preconditions made by a domain designer that preferred correct and readable descriptions.
In situations in which Task2 is never reached from the top-level tasks, it becomes a loose branch.
If reachable, but with the related Method4 containing a rigid precondition not declared in the initial state, it also becomes impossible, a dead branch.
If Task2 is a top-level task, the planning instance is considered impossible to solve before planning.
In any case Method5 may also become unreachable, which could lead to other removals.
One such case could be Method3 containing Task2 as a subtask, which could make more preconditions of Method2 to be pulled, as there is no other way to decompose such path.
The process repeats to deal with cycles within the domain, until a stable description is found.

\begin{figure}[ht]
\centering
\includegraphics[width=0.3\textwidth]{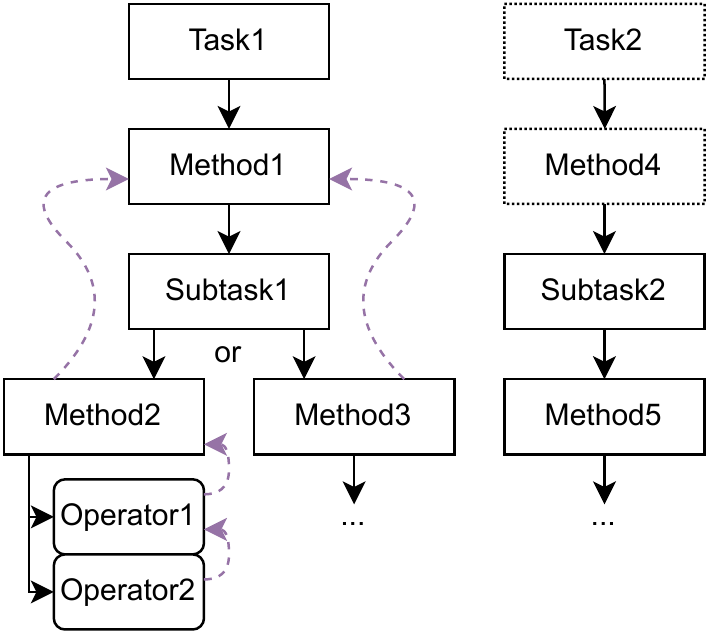}
\caption{The HTN domain may contain locally rigid, shared preconditions, loose and dead branches that are pulled up or removed by Pullup. Dotted lines mark the elements affected.}
\label{fig:pullup}
\end{figure}

Our implementation of Pullup follows Algorithm~\ref{alg:pullup}, with 3 main phases.
The first and final phases are used to remove loose and dead branches, the middle phase to move preconditions upwards.
Loose branches are branches that are never decomposed by the provided top-level tasks, or are impossible to be decomposed, as no value can replace a free variable.
Dead branches are branches that appear to be reachable, as they are connected to the top-level tasks, but are actually impossible once preconditions are evaluated.
Free variables that can only accept one value are replaced at the first and last steps, which are improved by the smaller sets of possible values created by the previous Typredicate extension.
A counter is used to discover how many ways an action or method can be reached.

In the pull up phase each method subtask is evaluated, and their locally rigid preconditions are moved up to the method.
Actions that can only be achieved by a single method, based on their counter, may have their preconditions completely pulled up.
Tasks that can be decomposed by multiple methods can only pull up preconditions shared by all methods.
To move more preconditions up we duplicate equality related preconditions to reveal hidden shared preconditions.
Multiple methods may have a precondition \textit{(at ?here)}, while another have \textit{(at ?there) (= ?here ?there)}.
The missing shared precondition is created to enable such hidden pull up.
The first phase only makes the following phases easier with less elements to consider.
The preconditions pulled up help the dead branch removal phase with the extra information unavailable during the first phase.
The Pullup extension is analogous to tree shaking/dead code elimination, code motion and common expression elimination in optimizing compilers~\cite{offutt1994using}.

\begin{algorithm}[!ht]
  \caption{Pullup}
  \label{alg:pullup}
  \begin{algorithmic}[1]
  \NoThen
\Procedure{Pullup}{\textbf{Op}, \textbf{M}, \textbf{I}, \textbf{T}}
      \State impossible = $\varnothing$
      \State counter = table of tasks, starting with 0
      \State increment counter for each \textbf{name}(\textbf{T})
      \For {m $\in$ \textbf{M}} \Comment{Remove loose branches}
        \For{pre $\in$ \textbf{pre}(m)}
          \State find applicable values for variables in pre
        \EndFor
        \IfThen{values = $\varnothing$}{remove m, \textbf{continue}}
        \For{v $\in$ values, $|$values$|$ = 1}
          \State replace v with fv $\in$ \textbf{variables}(m), remove fv
          \State replace v with fv $\in$ \textbf{pre}(m)
          \State replace v with fv $\in$ \textbf{tasks}(m)
        \EndFor
        \State increment counter for each \textbf{tasks}(m)
      \EndFor
      \For{o $\in$ \textbf{Op}, counter[o] = 0 $\land$ rigid \textbf{pre$^+$}(o) $\notin$ \textbf{I}}
        \State mark o as impossible and remove from \textbf{Op}
      \EndFor
      \State repeat = true \Comment{Pull preconditions up}
      \While{repeat}
        \State repeat = false
        \For{m $\in$ \textbf{M}}
          \State effects = table
          \For{s $\in$ \textbf{tasks}(m)}
            \If{s $\in$ impossible}
              \State repeat = true
              \State decrement counter for each \textbf{tasks}(m)
              \State remove operator o if counter[o] = 0
              \State remove m from \textbf{M}
              \State \textbf{break}
            \ElsIf{s $\in$ \textbf{Op}}
              \State p = \textbf{pre}(s) with variables from m
              \State copy p to \textbf{pre}(m)
              \State mark \textbf{eff}(s) in effects
              \IfThen{counter[s] = 1}{\textbf{pre}(s) = $\varnothing$}
            \Else
              \State m2 = \Call{decomposition}{s, \textbf{M}}
              \State prec = preconditions shared by m2
              \State prec = select prec without free variables 
              \State prec = select unmarked effects of prec
              \State \textbf{pre}(m) = \textbf{pre}(m) $\cdot$ prec
              \State traverse m2 and mark effects
            \EndIf
            \State eq = equality duplicates from \textbf{pre}(s)
            \State \textbf{pre}(m) = \textbf{pre}(m) $\cdot$ eq
          \EndFor
          \IfThen{\textbf{pre}(m) modified}{repeat = true}
        \EndFor
      \EndWhile
      \State Repeat removal step \Comment{Remove dead branches}
    \EndProcedure
  \end{algorithmic}
\end{algorithm}

\subsection{Dejavu}

Some domains may have methods with direct recursion, where a method includes the same task that it decomposes, or indirect recursion, requiring further decomposition before the (same) task is encountered. 
Without ``visited’’ predicates used by a domain expert to mark (register) and query visited partial states, such domains can induce an infinite loop for a TFD~\cite{ghallab2004automated} search procedure.
Dejavu transforms the domain by adding ``unobservable’’ primitive tasks (that are not part of valid plans) to mark and unmark the fact that a particular non-primitive task is being decomposed, and predicates to detect when the task is being recursively (re)attempted.
Information relating to such cycles is stored across decomposition branches using an external cache structure, as the state loses the marked information upon backtracking.
The cache saves which methods and unifications have been explored in previous branches to avoid repeating decompositions that previously led to failure. Domains with cyclic tasks without parameters lack the required information to cache the task signature, which contains the variable bindings for the method decomposing the task.
In such domains we fallback to a full state comparison with previously visited states at each cyclic task.
HyperTensioN can still detect stack overflows, and safely backtrack in case the cycle detection mechanism fails. 
Dejavu, while limited,  proved critical for HyperTensioN's performance, as it allows TFD to efficiently drive search, while avoiding its key limitation in recursive domains.

Dejavu, Algorithm~\ref{alg:dejavu}, starts detecting knots (points that may be revisited in cyclic tasks) by visiting each task from the provided top-level tasks.
Each knot is made of a task, method and decomposition.
Each knot focuses on a task that may decompose itself, within the subtasks at a certain index with certain terms having been applied before it.
The knot is considered useful based on empirical testing, it must match one of the conditions: method and task have the same name and decomposition has more than one subtask; decomposition has no free variables; task has no parameters; or task parameters are different from the applied terms.
If the condition is satisfied a new visited negative precondition is added to the decomposition and the guard visit/unvisit operator added around the cyclic subtask.
Create visit/unvisit operators if required.
Any visit/unvisit operator in the subtasks that happens before this decomposition is turned into a mark/unmark operator, a weaker cycle detector which does not store information in the cache, only in the state. 

\begin{algorithm}[ht]
  \caption{Dejavu}
  \label{alg:dejavu}
  \begin{algorithmic}[1]
  \NoThen
\Procedure{Dejavu}{\textbf{Op}, \textbf{M}, \textbf{T}}
      \IfThen {\textbf{T} = $\varnothing$}{\Return}
      \State knots = $\varnothing$
      \For {$t$ $\in$ \Call{unique}{T}}
        \State \Call{visit}{t, M, knots, $\varnothing$}
      \EndFor
      \For{$\langle$t, m, d$\rangle$ $\in$ knots}
        \State index = \Call{Find}{t, \textbf{tasks}(m)}
        \State terms = \textbf{terms}(\textbf{tasks}(m)[from 0 to index - 1])
        \If{
          \textbf{name}(m) = \textbf{name}(t) $\land$ $|$\textbf{tasks}(d)$|$ $>$ 1 $\lor$
          \textbf{variables}(d) = $\varnothing$ $\lor$
          \textbf{terms}(t) = $\varnothing$ $\lor$
          \textbf{terms}(t) $\neq$ terms
        }
          \State \Call{insert}{\textbf{pre$^-$}(d), $\langle$visited-knot, terms$\rangle$}
          \State \Call{insert-at}{\textbf{tasks}(d), index, $\langle$visit, terms$\rangle$}
          \State \Call{insert-at}{\textbf{tasks}(d), index+2, $\langle$unvisit, terms$\rangle$}
          \If{visit-knot $\notin$ \textbf{Op}}
\State \Call{insert}{\textbf{Op}, $\langle$visit-knot, terms$\rangle$}
            \State \Call{insert}{\textbf{Op}, $\langle$unvisit-knot, terms$\rangle$}
          \EndIf
          \For{dec $\in$ \Call{decomposition}{m, \textbf{M}}}
            \IfThen{dec = d}{break}
            \For{$\langle$subtask, sterms$\rangle$ $\in$ dec}
              \If{subtask = (un)visit $\land$ sterms = terms}
                \State \Call{rename}{subtask, (un)visit, (un)mark}
              \EndIf
            \EndFor
          \EndFor
        \EndIf
      \EndFor
    \EndProcedure

    \Procedure{Visit}{task, \textbf{M}, knots, visited}
      \If{task $\in$ \textbf{M}}
        \State visited = \Call{copy}{visited}
        \State \Call{Insert}{visited, \textbf{name}(task)}
        \For{d $\in$ \Call{decompositions}{task, \textbf{M}}}
          \For{t $\in$ \textbf{tasks}(d)}
            \If{\textbf{name}(t) $\in$ visited}
              \State \Call{insert}{knots, $\langle$t, m, d$\rangle$}
            \EndIf
\State \textbf{else} \Call{visit}{\textbf{name}(task), \textbf{M}, knots, visited}
          \EndFor
        \EndFor
      \EndIf
    \EndProcedure
  \end{algorithmic}
\end{algorithm}

\section{Comparison}

We now compare the improvements obtained by the above extensions w.r.t. some of the
IPC 2020 domains.
We selected 4 domains to comment the impact of the improvements.
The experiments used HyperTensioN 56be727~\footnote{https://github.com/Maumagnaguagno/HyperTensioN} and took place on Manjaro 21.2.6 with Ruby 3.0.4, Ryzen 3600XT with 16GB of RAM at 3200MT.
The front-end was improved by a C tokenizer extension named Ichor~\footnote{https://github.com/Maumagnaguagno/Ichor}.
The experiments used a 60s time-out and all combinations of the previously described extensions and the Ruby back-end used during the competition.
The Ruby VM stack has been set to 15MB, combined with the system stack set to 4GB using ulimit.
This allows deep plans to be completed, but may affect plan time and size in small instances without cycle detection.

\subsection{Woodworking}

\begin{figure}[!ht]
\centering
\includegraphics[width=0.35\textwidth]{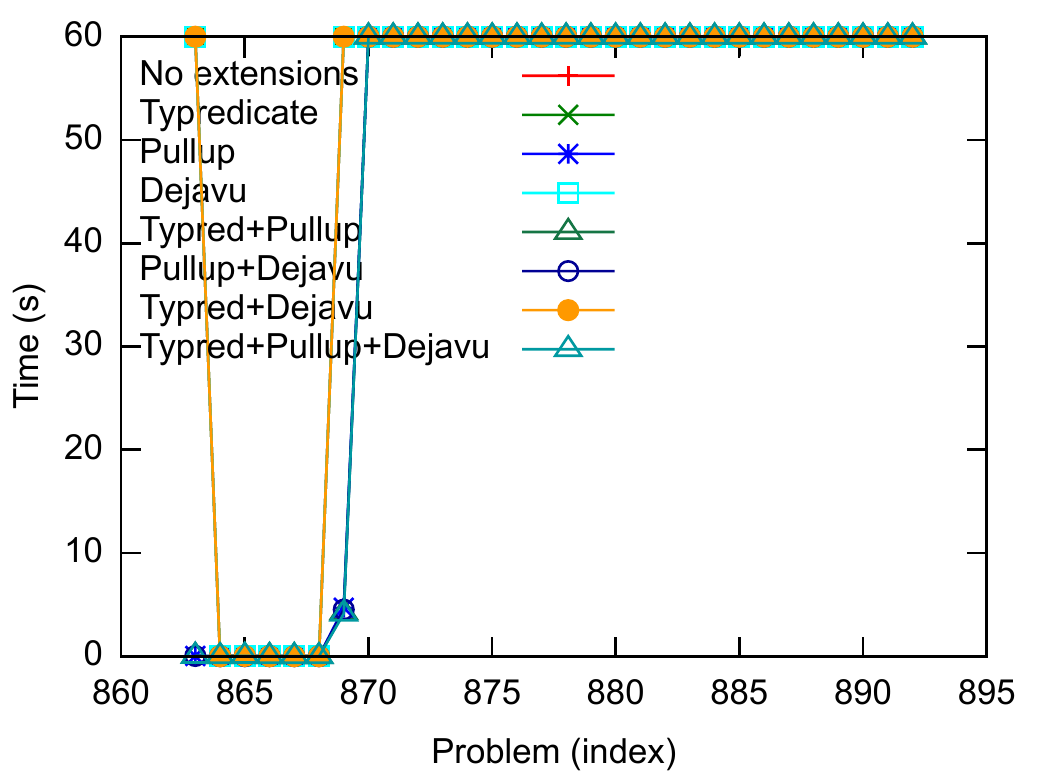}
\caption{Time in seconds to solve Woodworking problems.}
\label{fig:woodworking_plot}
\end{figure}

Woodworking~\cite{bercher2014hybrid} is based on a benchmark from earlier IPCs. It describes tasks for working with wood, such as cutting, polishing and finishing. With Pullup, two extra problems were solved within our time limit, with one of them taking less than a second as shown in Fig.~\ref{fig:woodworking_plot}.
Many problems in this domain seemed to require selecting the right values among many available objects before continuing exploration, as otherwise too much time was spent on backtracking, causing time-out.

\subsection{Rover-GTOHP}

\begin{figure}[!ht]
\centering
\includegraphics[width=0.35\textwidth]{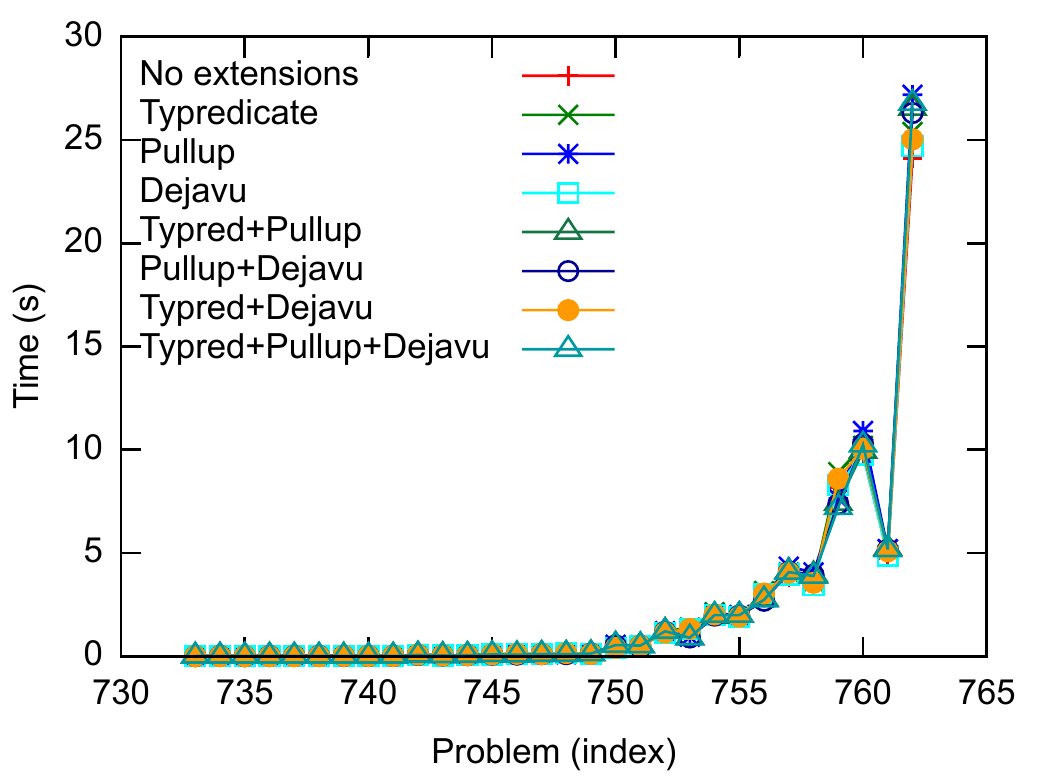}
\caption{Time in seconds to solve Rover-GTOHP problems.}
\label{fig:rover_plot}
\end{figure}

Rover involves robots navigating a planet, collecting information and sending it to a lander.
The HTN IPC domain~\cite{gtohp2017} is very similar to the one developed for SHOP~\cite{nau1999shop} based on problem instances from  earlier IPCs.
Differently from the original Rover used for testing prior to the competition, the extensions are closer to an overhead than optimizations, as shown in Fig.~\ref{fig:rover_plot}.
This suggests good usage of preconditions and visit/unvisit operators not needed or already added by the domain designer, which is the case in this domain.

\vspace{-10pt}

\subsection{Transport}

\begin{figure}[!ht]
\centering
\includegraphics[width=0.35\textwidth]{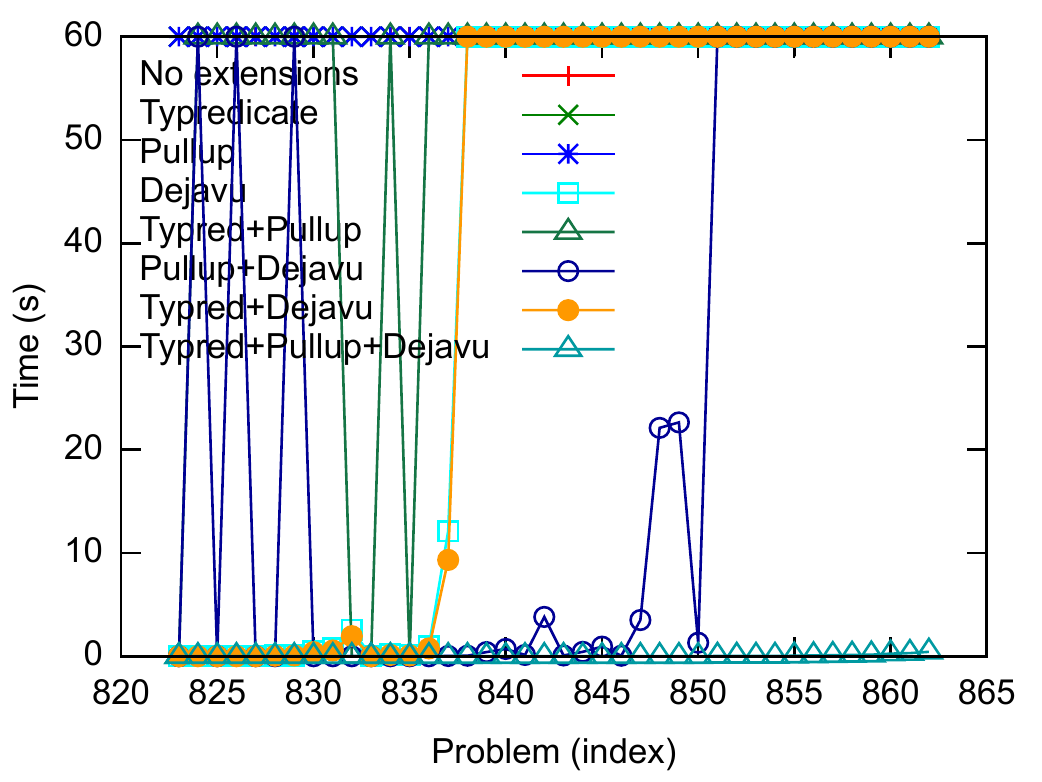}
\caption{Time in seconds to solve Transport problems.}
\label{fig:transport_plot}
\end{figure}

Transport~\cite{behnke2018totsat} describes a domain where delivery trucks with limited capacity must pick and drop packages at specific cities connected by a road network.
Transport is one of the few domains where each extension shows an impact on planning time, as shown in Fig.~\ref{fig:transport_plot}.
Typredicate is able to specialize the ``at’’ predicate, avoiding some unifications with non-vehicle objects. 
Pullup is able to move important constraints only defined in the leaves of the HTN structure, e.g. the need for a road between two cities in order to drive between them.
Note that this domain does not contain method preconditions.
With Typredicate and Pullup combined the Transport instances are easier to be solved, although they still require Dejavu for a greater optimization to take place, as many loops may happen.
Previous results always had the direct recursion detection of Dejavu enabled, in the current version we can see the actual impact of the cycle detection completely enabled and disabled.

\vspace{-10pt}

\subsection{Snake}

\begin{figure}[!ht]
\centering
\includegraphics[width=0.35\textwidth]{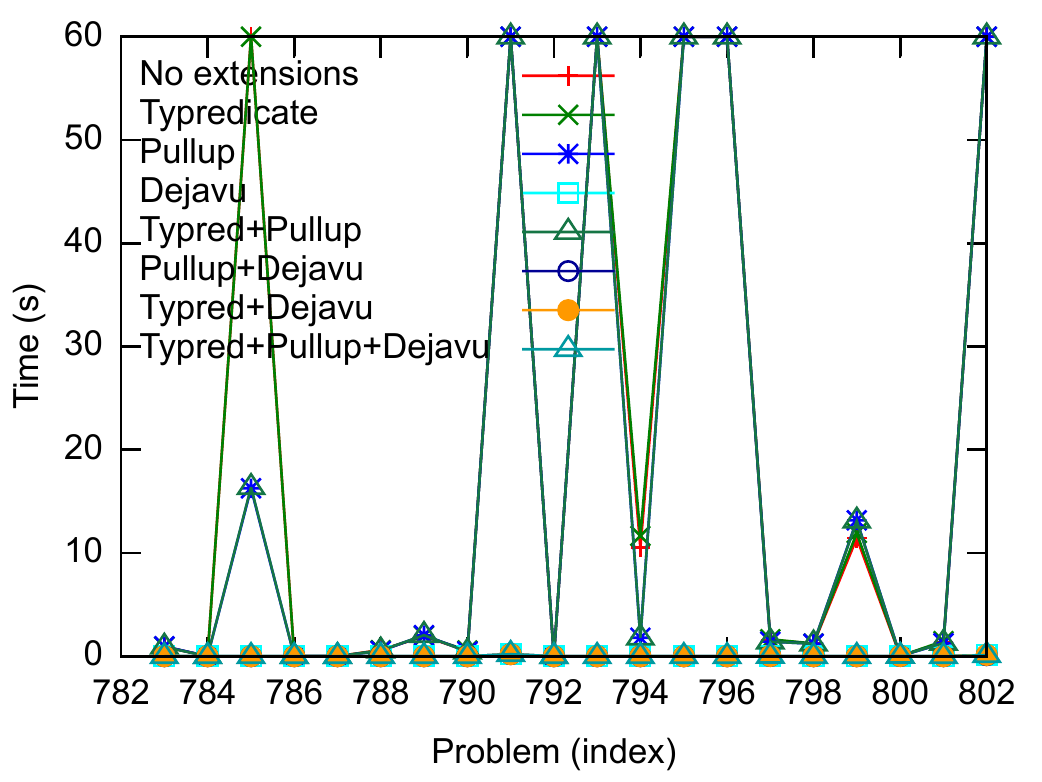}
\caption{Time in seconds to solve Snake problems.}
\label{fig:snake_plot}
\end{figure}

In Snake,\footnote{https://github.com/Maumagnaguagno/Snake} one or more snakes need to move to clear locations or strike nearby mice in a grid/graph-based world.
The domain benefits from Dejavu, i.e., the planner avoids unifications that recursively expands the same task, which may start an infinite loop.
Due to our stack overflow backtracking mechanism, this domain performs much worse with large stack size and Dejavu disabled, taking much longer to backtrack.
A better result can be obtained using the default smaller stack, or enabling Dejavu.
Since Dejavu's direct recursion detection is now controlled by the extension, its effect is visible in the graph, being required to avoid reaching the same positions repeatedly.
Observe from Fig.~\ref{fig:snake_plot} that Pullup shows a bigger improvement in planning time in the most complex instances.

\subsection{Results}

The experiments results are shown in Table~\ref{tab:ipc} with the amount of instances solved by each configuration for each domain, the highest values (sometimes obtained by multiple configurations) in bold.
Only two domains have no instances solved within the time restriction, Freecell-Learned and Monroe-PO.
Seven domains are unaffected by the extensions, such as Barman-BDI and Elevator-Learned.
Typredicate results are similar to No Extensions, the same happens for Pullup and Typredicate + Pullup, and Dejavu and Typredicate + Dejavu.
Typredicate is only useful for one domain, Transport, when combined with the other optimizations, being able to solve all instances.
Note that some extensions may add a significant overhead, such as Pullup in Factories simple, being worse than No extension.

Note that the IPC release of HyperTensioN was not able to parse the Entertainment and Monroe domains correctly.
With the parser fixed and all optimizations enabled, new results were obtained, which does not match the IPC samples' timings.
The first 5 of 12 Entertainment instances were solved in under 1s, the sixth in 7s, and the eighth in 44s; others exceeded the time limit (60s).
All Monroe-FO instances were solvable, most under 2 seconds and the last two in 4s.
The Monroe-PO instances were not solvable within the time limit.

\setlength{\tabcolsep}{2pt} 

\begin{table*}
    \small
    \centering
    \caption{HyperTensioN's multiple configurations.}
    \label{tab:ipc}
    \begin{tabular}{l|l|l|l|l|l|l|l|l}
        Domain(instances)        & No Ext       & Typred      & Pullup      & Dejavu      & Typred+Pullup & Pullup+Dejavu & Typred+Dejavu & Typred+Pullup+Dejavu \\ \hline
        AssemblyHierarchical(30) & 0            & 0           & 1           & 2           & 1             & \textbf{3}    & 2             & \textbf{3}           \\
        Barman-BDI(20)           & \textbf{20}  & \textbf{20} & \textbf{20} & \textbf{20} & \textbf{20}   & \textbf{20}   & \textbf{20}   & \textbf{20}          \\
        Blocksworld-GTOHP(30)    & \textbf{14}  & \textbf{14} & \textbf{14} & \textbf{14} & \textbf{14}   & \textbf{14}   & \textbf{14}   & \textbf{14}          \\
        Blocksworld-HPDDL(30)    & 0            & 0           & 0           & \textbf{28} & 0             & \textbf{28}   & \textbf{28}   & \textbf{28}          \\
        Childsnack(30)           & 27           & 27          & \textbf{30} & 27          & \textbf{30}   & \textbf{30}   & 27            & \textbf{30}          \\
        Depots(30)               & \textbf{23}  & \textbf{23} & \textbf{23} & \textbf{23} & \textbf{23}   & \textbf{23}   & \textbf{23}   & \textbf{23}          \\
        Elevator-Learned(147)    & \textbf{147} & \textbf{147}& \textbf{147}& \textbf{147}& \textbf{147}  & \textbf{147}  & \textbf{147}  & \textbf{147}         \\
        Entertainment(12)        & 3            & 3           & 3           & 5           & 3             & \textbf{7}    & 5             & \textbf{7}           \\
        Factories-simple(20)     & 1            & 1           & 0           & \textbf{3}  & 0             & \textbf{3}    & \textbf{3}    & \textbf{3}           \\
        Freecell-Learned(60)     & 0            & 0           & 0           & 0           & 0             & 0             & 0             & 0                    \\
        Hiking(30)               & 0            & 0           & 0           & \textbf{25} & 0             & \textbf{25}   & \textbf{25}   & \textbf{25}          \\
        Logistics-Learned(80)    & 0            & 0           & 0           & \textbf{22} & 0             & \textbf{22}   & \textbf{22}   & \textbf{22}          \\
        Minecraft-Player(20)     & 3            & 3           & 3           & \textbf{5}  & 3             & \textbf{5}    & \textbf{5}    & \textbf{5}           \\
        Minecraft-Regular(59)    & \textbf{56}  & \textbf{56} & \textbf{56} & \textbf{56} & \textbf{56}   & \textbf{56}   & \textbf{56}   & \textbf{56}          \\
        Monroe-FO(20)            & 6            & 6           & \textbf{20} & 6           & \textbf{20}   & \textbf{20}   & 6             & \textbf{20}          \\
        Monroe-PO(20)            & 0            & 0           & 0           & 0           & 0             & 0             & 0             & 0                    \\
        Multiarm-Blocksworld(74) & 2            & 2           & 3           & \textbf{8}  & 3             & \textbf{8}    & \textbf{8}    & \textbf{8}           \\
        Robot(20)                & 6            & 6           & 6           & \textbf{14} & 6             & \textbf{14}   & \textbf{14}   & \textbf{14}          \\
        Rover-GTOHP(30)          & \textbf{30}  & \textbf{30} & \textbf{30} & \textbf{30} & \textbf{30}   & \textbf{30}   & \textbf{30}   & \textbf{30}          \\
        Satellite-GTOHP(20)      & 0            & 0           & 0           & 3           & 0             & \textbf{20}   & 3             & \textbf{20}          \\
        Snake(20)                & 14           & 14          & 15          & \textbf{20} & 15            & \textbf{20}   & \textbf{20}   & \textbf{20}          \\
        Towers(20)               & \textbf{13}  & \textbf{13} & \textbf{13} & \textbf{13} & \textbf{13}   & \textbf{13}   & \textbf{13}   & \textbf{13}          \\
        Transport(40)            & 0            & 0           & 0           & 15          & 4             & 25            & 15            & \textbf{40}          \\
        Woodworking(30)          & 5            & 5           & \textbf{7}  & 5           & \textbf{7}    & \textbf{7}    & 5             & \textbf{7}           \\ \hline
        \textbf{Total(892)}      & 370          & 370         & 391         & 491         & 395           & 540           & 491           & \textbf{555}         \\       
\end{tabular}
\end{table*}

\section{Conclusion}

This paper presented HyperTensioN, an approach to planning using a three-stage compiler designed to support optimizations in multiple domain description languages. The flexibility introduced by the front and back-end modules makes it easy to support new domain description languages, while the middle-end pipeline opens the door for multiple transformation and analysis tools to be executed before planning.
The key to its performance in the IPC is a set of domain transformation techniques that replicates domain-knowledge optimizations commonly used to speed up search in previous HTN planners such as JSHOP2~\cite{Ilghami2003}, as well as the optimizations often used by agent interpreters, e.g.~\cite{Thangarajah2003a}. With our domain transformations it was possible to not only improve the HTN structure for SHOP-like (blind Depth First-Search) planners using Typredicate and Pullup, but also to avoid recomputing parts of complex combinatoric domains such as Transport and Snake using Dejavu. Future work includes stronger tree modifications/specializations, support for more complex domain descriptions, and heuristic functions to improve search.
\\
\textbf{Acknowledgements:} Felipe Meneguzzi acknowledges support from CNPq with projects 407058/2018-4 (Universal) and 302773/2019-3 (PQ Fellowship).

\end{document}